\documentclass[10pt,twocolumn,letterpaper]{article}

\usepackage[pagenumbers]{cvpr} 

\usepackage{graphicx}
\usepackage{amsmath}
\usepackage{amssymb}
\usepackage{booktabs}
\usepackage{bm}

\usepackage[pagebackref,breaklinks,colorlinks]{hyperref}

\usepackage[capitalize]{cleveref}
\crefname{section}{Sec.}{Secs.}
\Crefname{section}{Section}{Sections}
\Crefname{table}{Table}{Tables}
\crefname{table}{Tab.}{Tabs.}

\usepackage{multirow}

\def\cvprPaperID{*****} 
\def\confName{CVPR}
\def\confYear{2022}

\begin{document}

\title{Sequential Transformer for End-to-End Person Search}

\author{Long Chen, Jinhua Xu\\
East China Normal University, Shanghai, China\\
{\tt\small longchen@stu.ecnu.edu.cn, jhxu@cs.ecnu.edu.cn}
}
\maketitle

\begin{abstract}
   Person Search aims to simultaneously localize and recognize a target person from realistic and uncropped gallery images.
   One major challenge of person search comes from the contradictory goals of the two sub-tasks, \textit{i.e.}, person detection focuses on finding the commonness of all persons so as to distinguish persons from the background, while person re-identification (re-ID) focuses on the differences among different persons. 
   In this paper, we propose a novel Sequential Transformer (SeqTR) for end-to-end person search to deal with this challenge.
   Our SeqTR contains a detection transformer and a novel re-ID transformer that sequentially addresses detection and re-ID tasks.
   The re-ID transformer comprises the self-attention layer that utilizes contextual information and the cross-attention layer that learns local fine-grained discriminative features of the human body. 
   Moreover, the re-ID transformer is shared and supervised by multi-scale features to improve the robustness of learned person representations. 
   Extensive experiments on two widely-used person search benchmarks, CUHK-SYSU and PRW, show that our proposed SeqTR not only outperforms all existing person search methods with a 59.3$\%$ mAP on PRW but also achieves comparable performance to the state-of-the-art results with an mAP of 94.8$\%$ on CUHK-SYSU.

\end{abstract}

\section{Introduction}
\label{sec:intro}

\begin{figure}[thpb]
  \setlength{\abovecaptionskip}{0.cm}
  \begin{center}
    \includegraphics[width=0.478\textwidth]{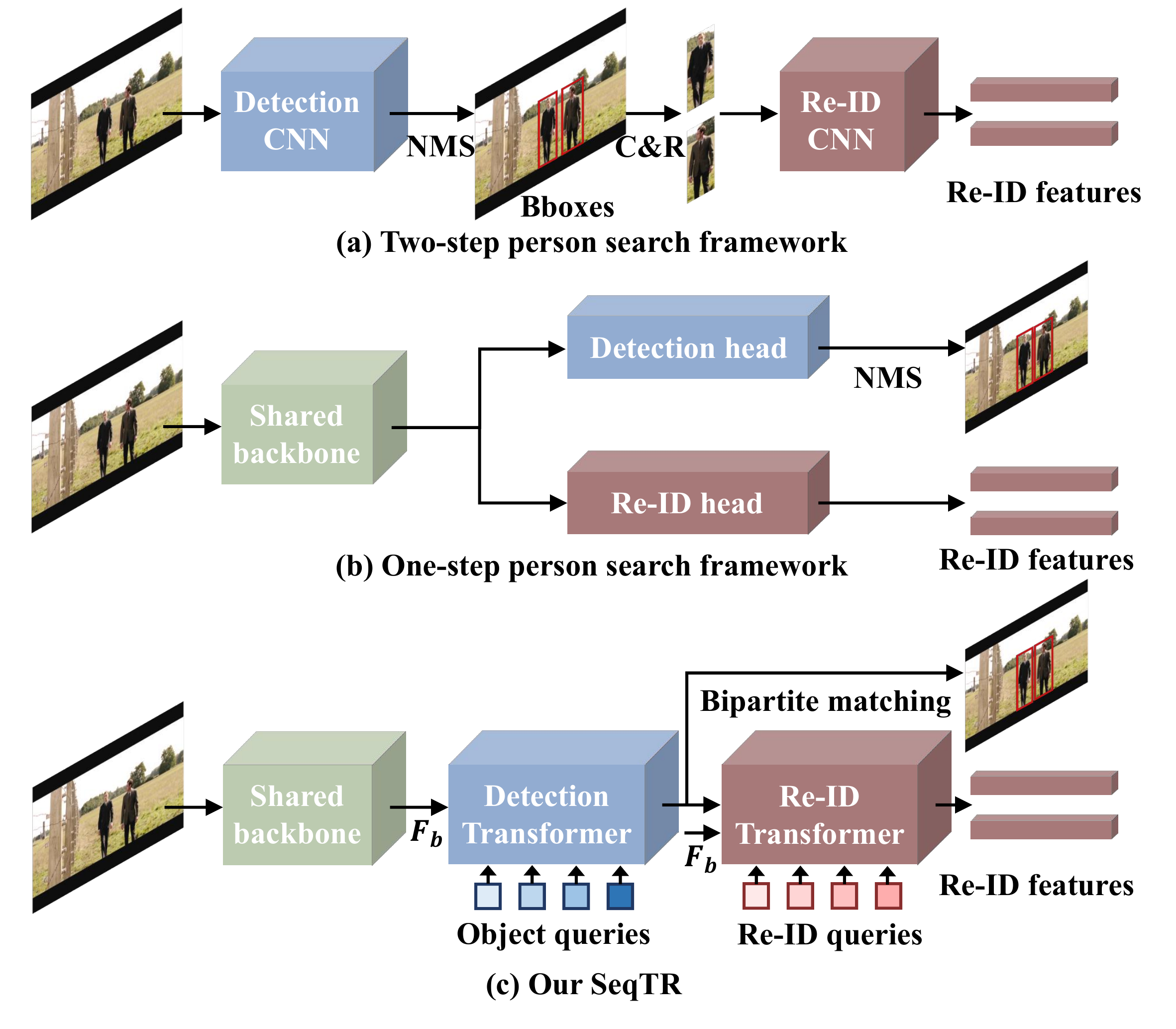}
    \caption{Comparison of person search frameworks. 
    (a) The  two-step  framework. 
    (b) The one-step  framework. 
    (c) Our proposed SeqTR adopts the sequential framework to perform detection and re-ID in order.}
    \label{fig:intro}
  \end{center}
  \vspace{-0.2cm}
\end{figure}

Practical applications of person search, such as searching for suspects and missing people in intelligent surveillance, require separating people from complex background and discriminating target identities (IDs) from other IDs. It involves two fundamental tasks in computer vision, \textit{i.e.}, pedestrian detection and person re-identification (re-ID). 
Pedestrian detection aims at detecting the bounding boxes (Bboxes) of all candidates in the image. Person re-ID aims at retrieving a person of interest across multiple non-overlapping cameras.
Person search has recently attracted tremendous interest of researchers in the computer vision community for its importance in building smart cities. However, it remains a difficult task that suffers from many challenges, such as jointly optimizing contradictory objectives of two sub-tasks in a unified framework, scale/pose variations, background clutter and occlusions and so on.

According to training manners, existing person search methods can be generally grouped into two categories: two-step frameworks and one-step frameworks.
Two-step methods typically perform detection and re-ID with two separate independent models.
As shown in Fig. \ref{fig:intro}(a), pedestrians are first detected by an off-the-shelf detection model. 
After non-maximum suppression (NMS), the person patches are cropped and resized (C\&R) into a fixed size. Then the person re-ID model is applied to produce ID feature embeddings, which will be used to calculate the similarity between the query persons and the candidates. The two-step frameworks can achieve satisfactory performance since each step focuses on one task and no contradictory is involved. However, this pipeline is time-consuming and resource-consuming. 
In contrast, one-step methods simultaneously optimize two sub-tasks in a joint framework (Fig. \ref{fig:intro}(b)). The two sub-tasks first share a common backbone for features extraction and then detection head and re-ID head are applied in parallel. 

In terms of architecture, the sequential framework combines the merits of two-step and one-step frameworks. It not only inherits the better performance of two-stage frameworks via providing accurate bounding boxes (Bboxes) for the re-ID stage but also preserves the efficiency of the end-to-end training manner of one-step frameworks. 
However, as Li \etal~ \cite{li2021sequential} has pointed out, the performance bottleneck of this architecture lies in the design of the re-ID sub-network. 
In addition, we find that NMS, commonly used in the detection models, primarily hinders the inference speed of this architecture, especially in crowded scenes. 

As transformers \cite{vaswani2017attention} become popular in vision tasks, transformers-based person search frameworks \cite{cao2022pstr, yu2022cascade} also show advantages over CNN-based models, such as no NMS needed and powerful capability of learning fine-grained features.  

Motivated by the above observations, we propose a novel Sequential transformer (SeqTR) for end-to-end Person Search (Fig. \ref{fig:intro}(c)). 
It is a sequential framework, in which two transformers are integrated seamlessly to address the detection and re-ID tasks. Meanwhile, the two transformers are decoupled with different features for the two contradictory tasks.  

In summary, we make the following contributions:
\begin{itemize}
  \item We propose a novel Sequential Transformer (SeqTR) model for end-to-end person search, which utilizes two transformers to sequentially perform pedestrian detection and re-ID without NMS post-processing.
  \item We propose a novel re-ID transformer to generate discriminative re-ID feature embeddings. To make full use of context information, we introduce the self-attention mechanism in our re-ID transformer. Meanwhile, we employ multiple cross-attention layers to learn local fine-grained features. To obtain scale-invariant person representations, our re-ID transformer is shared by multi-scale features.
  \item We achieve a state-of-the-art result on two datasets. Comprehensive experiments show the merits of our proposed modules. Furthermore, with PVTv2-B2  \cite{wang2022pvt} backbone, SeqTR achieves 59.3\% mAP that outperforms all existing person search models on PRW  \cite{zheng2017person}.
\end{itemize}

\begin{figure*}[t!]
  \setlength{\abovecaptionskip}{0.cm}
  \begin{center}
    \includegraphics[width=1\textwidth]{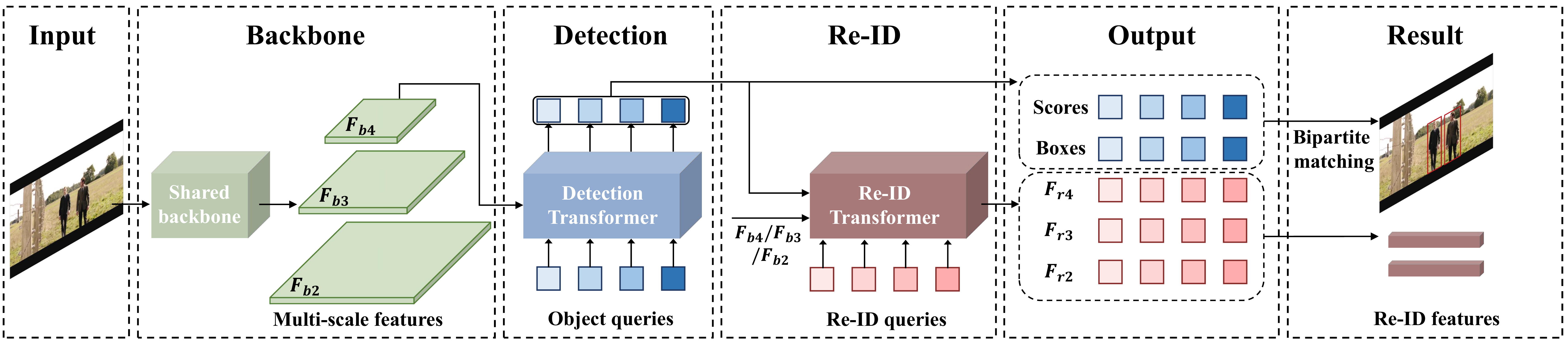}
    \caption{Architecture of our proposed SeqTR, which comprises a backbone, a detection transformer and a re-ID transformer.}
    \label{framework}
  \end{center}
  \vspace{-0.3cm}
\end{figure*}

\section{Related Work}
\label{sec:related}

\subsection{CNN-based Person Search}

Person search has attracted a lot of attention from the computer vision community. A large number of methods have been proposed and achieved remarkable results.
According to the training manner, existing person search frameworks can be divided into two-step and one-step methods.
Two-step person search models first perform pedestrian detection and subsequently crop the detected people for re-ID. 
Zheng \etal~ \cite{zheng2017person} first exhaustively evaluate the combinations of different detectors and re-ID models.
Chen \etal~ \cite{chen2018person} propose a mask-guided two-stream network to obtain enhanced feature representation. 
Lan \etal~\cite{lan2018person} analyze the multi-scale misalignment caused by the detector and exploit knowledge distillation to address it.
Wang \etal~\cite{wang2020tcts} utilize an identity-guided query detector to extract the query-like
proposals and employ a detection-adapted model for re-ID.
One-step person search models integrate detection and re-ID into a joint framework, which enables end-to-end training of two sub-tasks.
Xiao \etal~\cite{xiao2017joint} propose the first one-step person search model by introducing a re-ID branch and Online Instance Matching (OIM) loss in the Faster R-CNN detector.
Liu \etal~\cite{liu2017neural} and Chang \textit{et al.}\cite{chang2018rcaa} discard the proposal generation operation and search the query person directly on the uncropped images by sequential decision making or reinforcement learning.
Xiao \etal~\cite{xiao2019ian} use Center Loss to enhance feature discrimination.
Yan \etal~\cite{yan2019learning} enrich the features with surrounding persons.
Munjal \etal~\cite{munjal2019query} build the relationship between the query image and gallery image by integrating a query-guided Siamese squeeze-and-excitation block into the backbone.
Han \etal~\cite{han2019re} develop an RoI transform layer that enables gradient flow from the re-identifier to the detector for localization refinement.
Chen \etal~\cite{chen2020norm} propose a norm-aware embedding (NAE) to improve re-ID performance.
Dong \etal~\cite{dong2020bi} employ a Siamese network that takes both the entire image and cropped persons to better guide the feature learning of the person.
Yan \etal~\cite{yan2021anchor} introduce the first anchor-free approach for person search.
Li \etal~\cite{li2021sequential} propose a Sequential End-to-end Network (SeqNet) to obtain accurate Bboxes for the re-ID stage, in which detection and re-ID are considered as a progressive process and tackled with two sub-networks sequentially. SeqNet inherits the sequential process of two-stage methods and the end-to-end training fashion and efficiency of the one-step methods. Our work is inspired by SeqNet, and we use the sequential framework and replace the CNN sub-networks for detection and re-ID with two transformers. Employing the structure advantage of the transformer, no NMS is needed during training and inference, and the two sub-networks are integrated with deformable attention seamlessly rather than the ROI-align in SeqNet.

\subsection{Transformer-based Person Search}
Recently, transformers-based person search frameworks \cite{cao2022pstr, yu2022cascade} have been also proposed.
The COAT model \cite{yu2022cascade} is a cascaded one-step method, in which an occluded attention transformer is used for feature enhancement before the parallel detection head and re-ID head. In PSTR \cite{cao2022pstr}, a detection decoder and a re-ID decoder are designed for the two tasks. The output features of the detection decoder are fed into the re-ID decoder, therefore the two decoders with contradictory goals are coupled.
Considering the advantages of the transformer, we aim to utilize the transformer to design a robust re-ID sub-network to alleviate the performance bottleneck of the sequential framework.

\section{Method}
\label{sec:method}

In this section, we introduce our proposed SeqTR in detail.
Firstly, we give an overview architecture of SeqTR in Sec. \ref{subsec:SeqTR Architecture}.
Secondly, the details of our designed re-ID transformer are elaborated in Sec. \ref{subsec:re-ID transformer}.
Finally, we introduce the training and inference process in Sec. \ref{subsec:training and inference}.

\subsection{SeqTR Architecture}
\label{subsec:SeqTR Architecture}

The overall architecture of our SeqTR is depicted in Fig. \ref{framework}. It contains three main components: a  backbone to extract multi-scale feature maps of the input image, a detection transformer to predict Bboxes, and a novel re-ID transformer to learn robust person feature embeddings.

\textbf{Backbone.}
Starting from the initial image $x_{img}\in \mathbb{R}^{3 \times H_0 \times W_0}$ (with 3 color channels). 
The backbone extracts original multi-scale feature maps $\{x^{l}\}_{l=1}^3$ from stages $P_2$ through $P_4$ in PVTv2-B2 \cite{wang2022pvt} (or from stages $C_3$ through $C_5$ in RestNet \cite{he2016deep}). The resolution of $x^l$ is $2^{l+2}$ lower than the input image.

\textbf{Detection Transformer.}
We introduce the transformer-based detector, deformable DETR \cite{zhu2020deformable}, into our framework to predict the pedestrian bounding boxes. 
However, The difference with the original deformable DETR is the input features. 
First, the channel dimensions of all feature maps $\{x^{l}\}_{l=1}^3$ from the backbone are mapped to a smaller dimension $d$ = 256 by 1$\times$1 convolution.
Then, a 3$\times$3 deformable convolution is used to generate more accurate feature maps. 
Finally, $\{F_{bi}\in \mathbb{R}^{d\times H\times W}\}_{i=2}^4$ are transfomed from original feature maps $\{x^{l}\}_{l=1}^3$ by the above two steps and fed into a standard deformable DETR.

\textbf{re-ID Transformer.}
Our re-ID transformer aims to adaptively learn discriminative re-ID features around the human body center. 
Motivated by object queries in DETR \cite{carion2020end}, we set a fixed number of learnable re-ID queries $Q_r$ to reconcile the relationship between detection and re-ID and obtain re-ID feature embeddings. 


\subsection{re-ID transformer}
\label{subsec:re-ID transformer}

\begin{figure}[t!]
  \setlength{\abovecaptionskip}{0.cm}
  \begin{center}
    \includegraphics[width=0.47\textwidth]{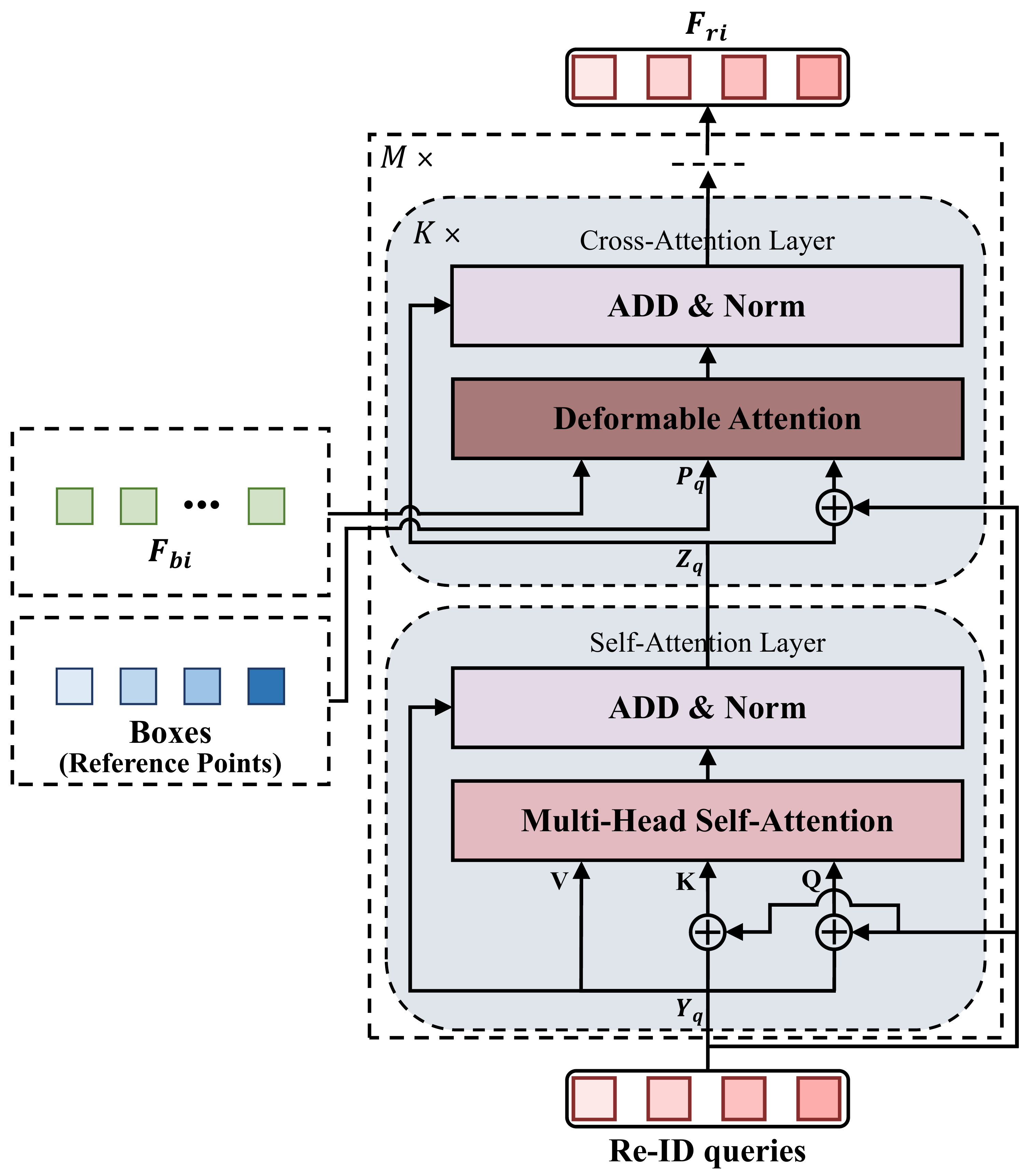}
    \caption{Architecture of our proposed re-ID transformer.
    }
    \label{fig:decoder}
  \end{center}
\end{figure}

The architecture of the re-ID transformer is shown in Fig. \ref{fig:decoder}.
Each re-ID transformer layer is composed of  a self-attention layer and $K$ cross-attention layers. The self-attention layer comprises a multi-head attention module and a layer normalization. The cross-attention layer contains a deformable attention module and a layer normalization.
Suppose that the detection transformer decodes $N$ objects in each image. The re-ID query number is also set as $N$. 
Taking the enhanced backbone features $F_{bi}, i \in [2, 4]$, $N$ reference points $P_q$ from the detection transformer and $N$ re-ID queries $Q_r$ as input, the  re-ID transformer outputs $N$ instance-level re-ID embeddings $F_{ri}$ that have the same dimension as the pixel features. 
These instance-level re-ID feature embeddings are highly associated with pedestrian locations. 
Furthermore, to aggregate multi-scale features, multi-scale feature maps $\{F_{bi}\}_{i=2}^4$ are  used to generate multi-scale re-ID embeddings $\{F_{ri}\in \mathbb{R}^{d\times H\times W}\}_{i=2}^4$ by the re-ID transformer.
During inference, all multi-scale re-ID embeddings $\{F_{ri}\}_{i=2}^4$ are concatenated to perform matching.


\textbf{Re-ID Queries.}
To  mitigate the objective contradictory problem, we set re-ID queries $Q_r$, like object queries, to obtain re-ID features.
Specifically, re-ID queries guarantees that the final re-ID embeddings $\{F_{ri}\}_{i=2}^4$ are  instance-level fine-grained features learned from the augmented multi-scale backbone features $\{F_{bi}\}_{i=2}^4$.
Through this design, the final learned re-ID feature embeddings are highly correlated with the detected pedestrian locations, but not affected by the detection features.
This is different from the re-ID decoder in PSTR \cite{cao2022pstr}, in which the re-ID queries come from the output features of the detection decoder.

\textbf{Self-Attention Layer.}
To produce discriminative re-ID feature embeddings, we introduce the self-attention layer into the re-ID transformer to learn contextual information.
This is different from the re-ID decoder in PSTR \cite{cao2022pstr}, in which no self-attention layer is used. From the ablation study (Table  \ref{Tab:SAL}) in experiments, the performance is improved with the self-attention layer.
Specifically, we adopt a standard multi-head self-attention (with $H$ heads) in the Transformer \cite{vaswani2017attention}. We denote the input of the self-attention layer as $Y_q$. 
The initial input $Y_q = Q_r$. $Y_q$  are transformed into query vectors $Q \in \mathbb{R}^{N \times d_k}$, key vectors $K \in \mathbb{R}^{N \times d_k}$ and value vectors $V \in \mathbb{R}^{N \times d_v}$ by three different linear projections. The output embeddings then are generated by performing the multi-head self-attention module. 
\begin{equation}
  {\rm head}_i = {\rm Attention} (QW^Q_i, KW^K_i, VW^V_i),
\end{equation}
where  $W^Q_i \in \mathbb{R}^{d\times d_k}$, $W^K_i\in \mathbb{R}^{d\times d_k}$, $W^V_i\in \mathbb{R}^{d\times d_v}$, $d_k = d_v = d/H$. The self-attention module use Scaled Dot-Product Attention in each head:
\begin{equation}
  {\rm Attention}(Q, K, V) = {\rm softmax} \left(\frac{QK^{\top}}{\sqrt{d_k}}\right)V.
  \label{eq:attention}
\end{equation}
The embeddings from all heads are concatenated and projected to yield $d$-demensional embeddings:
\begin{equation}
  {\rm MultiHead}(Q, K, V) = {\rm Concat} ({\rm head}_1, ..., {\rm head}_H) W^O,
\label{eq:MHA}
\end{equation}
where $W^O \in \mathbb{R}^{Hd_k\times d}$.  At last, we use a layer normalization to get the final embeddings $\hat{Y}_q$.
\begin{equation}
  \hat{Y}_q = {\rm layernorm} (Y_q + {\rm dropout} ({\rm MultiHead}(Q, K, V)).
\label{eq:layernorm}
\end{equation}

The self-attention layer in the first re-ID transformer layer can be skipped.
 After passing through the first re-ID transformer layer, the output features are correlated with reference points. $N$ feature embeddings correspond to $N$ locations respectively.
These embeddings  interact with each other for learning spatial relationship by the self-attention layer in the $m^{th}$ ($m \in [2, M]$) re-ID transformer layer, resulting to enhance feature embeddings by instances in the same scene.

\textbf{Cross-Attention Layer.}
Different from the previous works that use the RoI-Align layer on detection features, we employ and stack several cross-attention layers to address the region misalignment. 
In the cross-attention layer, there is a deformable attention module and a layer normalization. The deformable attention module proposed by deformable DETR \cite{zhu2020deformable}, only attends to a small set of key sampling points around a reference point. It is useful for learning fine-grained features. 
Given an input feature map $F_{bi}\in \mathbb{R}^{C\times H\times W}$, a set of detected bounding boxes, \textit{i.e.}, reference points (denoted $P_q$), and query features (denoted $Z_q$), the output feature embeddings $\hat{Z}_q$ can be calculated:
\begin{equation}
  \begin{split}
    \hat{Z}_q &= \\
    &{\rm layernorm} (Z_q + {\rm dropout} ({\rm DeformAttn}(Z_q, P_q,F_{bi})),
   \label{eq:CAlayernorm}
  \end{split}
\end{equation}
\begin{equation}
  \begin{split}
    {\rm DeformAttn}(Z_q&, P_{q}, F_{bi}) =\\
    &\sum_{h=1}^{H}W_h \left[\sum_{s=1}^{S}A_{hs} \cdot W_h^\prime F_{bi}(P_q + \Delta P_{hs}) \right]
   \label{eq:DA}
  \end{split}
\end{equation}
where $H$ is the total attention heads, $S$ is the total sampled key number. $A_{hs}$ and $\Delta P_{hs}$ denote attention weight of the $s^{th}$ sampling point in the $h^{th}$ attention head and the sampling offset, respectively. 
Both are obtained via linear projection over the query feature $Z_q$, respectively. In this way, each  query feature corresponds to one detected bounding boxes and integrates the features of the surround sampling points.
In PSTR \cite{cao2022pstr}, features at sampling points are averaged rather than using the attention weight $A_{hs}$ as in Eq. \ref{eq:DA} because it was observed that the attention weights from the query struggle to effectively capture the features of a person instance. 
We think it may be caused by the coupling of the two decoders since the re-ID queries in PSTR  \cite{cao2022pstr} are from the detection decoder.

\begin{figure}[thpb]
  \setlength{\abovecaptionskip}{0.cm}
  \begin{center}
    \includegraphics[width=0.478\textwidth]{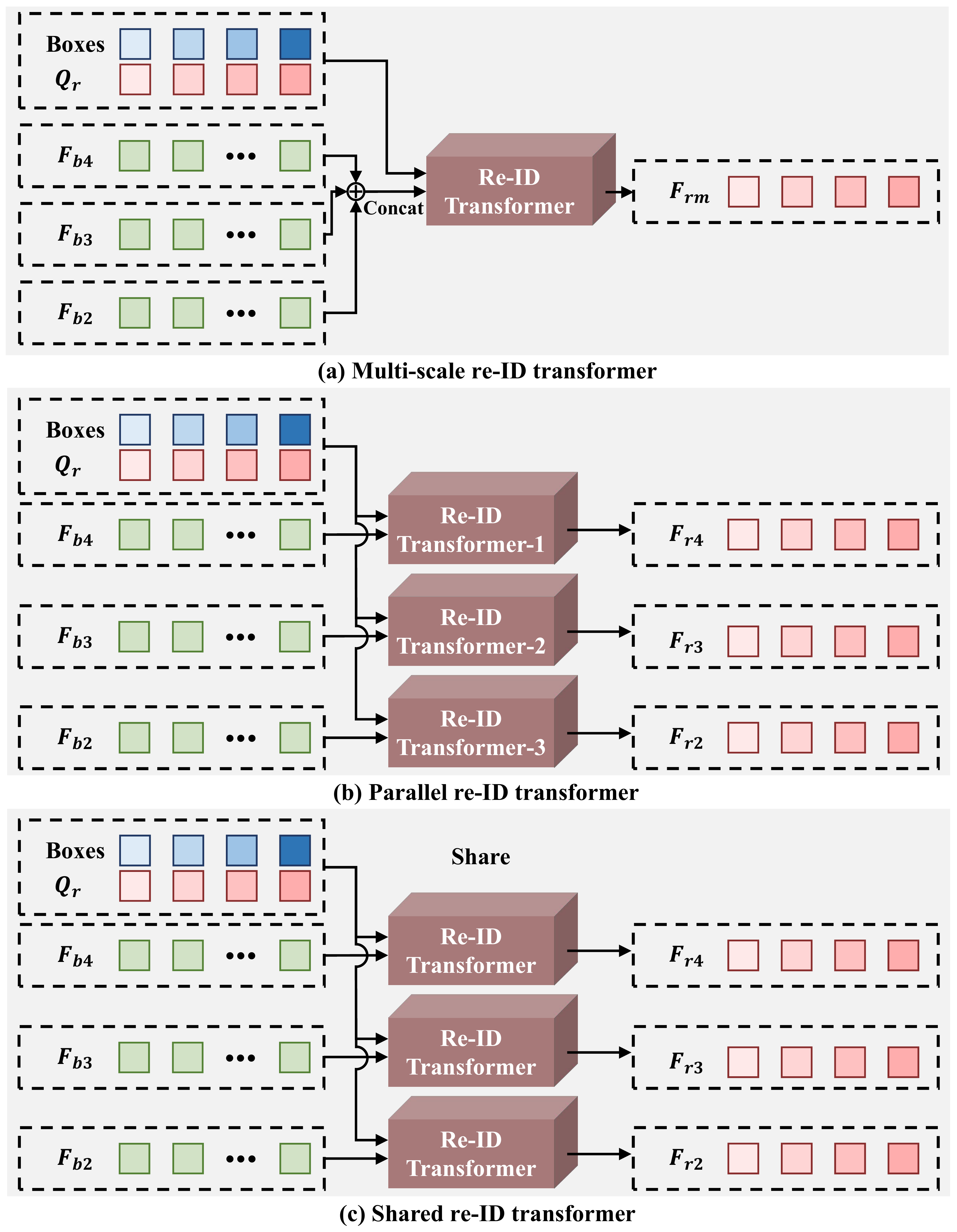}
    \caption{Comparison of different re-ID transformer schemes.
    (a)  Multi-scale re-ID transformer. 
    $\{F_{bi}\}_{i=2}^4$ are concatenated as input features.
    (b) Parallel re-ID transformer. Each independent re-ID transformer is  responsible for a single-scale input feature.
    (c) Shared re-ID transformer. $\{F_{bi}\}_{i=2}^4$ respectively go through a common shared re-ID transformer.}
    \label{fig:RDDS}
  \end{center}
  \vspace{-0.3cm}
\end{figure}

\textbf{Schemes of Employing Multi-scale Features.}
Much previous work has demonstrated that employing multi-scale feature maps is useful for addressing scale variation in person search.
To obtain scale-invariant re-ID features, we propose several schemes of employing multi-scale features.
First, A straightforward way is to concatenate the augmented backbone features $\{F_{bi}\}_{i=2}^4$ and feed to the re-ID transformer to produce $F_{rm}\in \mathbb{R}^{N\times d}$, as shown in Fig. \ref{fig:RDDS}(a). To align the dimension of the final matching embeddings with other schemes in Fig. \ref{fig:RDDS}, we also design "Multi-scale re-ID transformer-$3d$", whose deformable attention modules are replaced by multi-scale deformable attention modules \cite{zhu2020deformable}. Correspondingly, the re-ID queries are adjusted to $Q_r\in \mathbb{R}^{N\times 3d}$, resulting in $F_{rm}\in \mathbb{R}^{N\times 3d}$. 
We also build three independent re-ID transformers for three-level features $\{F_{bi}\}_{i=2}^4$ to obtain three-level re-ID feature embeddings $\{F_{ri}\in \mathbb{R}^{N\times d}\}_{i=2}^4$, respectively. we call it parallel re-ID transformer (Fig. \ref{fig:RDDS}(b)).
As opposed to parallel re-ID transformer, the shared re-ID transformer (Fig. \ref{fig:RDDS}(c)) means that three-scale input features are respectively fed to a common re-ID transformer to generate re-ID feature embeddings. The following ablation studies (Table \ref{Tab:DRD}) verify that the shared re-ID transformer achieves the best performance. 

\subsection{Training and Inference}
\label{subsec:training and inference}

For each image, our SeqTR predicts $N$ classification scores, bounding boxes and re-ID feature embeddings $\{F_{ri}\}_{i=2}^4$. 
In the training phase, $\{F_{ri}\}_{i=2}^4$ are supervised separately. They are concatenated  during inference. 

During training, our SeqTR is trained end-to-end for detection and re-ID. Specifically, detection transformer is supervised with loss functions of deformable DETR \cite{zhu2020deformable} for classification ($L_{cls}$), bounding-box IoU loss ($L_{iou}$), bounding-box Smooth-L1 loss ($L_{cls}$). While the re-ID transformer is supervised by the Focal OIM loss ($L_{oim}$) \cite{yan2021anchor}.

The overall loss is given by:
\begin{equation}
  \begin{split}
    L = \lambda_{1}L_{cls} + \lambda_{2}L_{iou} + \lambda_{3}L_{l1} + \lambda_{4}L_{oim}
    \label{eq:loss}
  \end{split}
\end{equation}
where $\lambda_{1}$,$\lambda_{2}$,$\lambda_{3}$,$\lambda_{4}$, responsible for the relative loss importance, are set as 2.0, 5.0, 2.0, 0.5, respectively.

During inference, our SeqTR predicts Bboxes and corresponding re-ID feature embeddings for gallery images. For the query person, we get predictions of the query image in the same way and then choose the one that has maximum overlap with its annotated bounding box.

\section{Experiments}
\label{}

In this section, we conduct experiments on two widely utilized person search datasets.
We first introduce two large datasets and evaluation metrics.
Then we describe some implementation details.
Afterwards, we compare the overall performance of our methods with state-of-the-art methods.
Finally, we perform ablation studies to validate the effectiveness of our methods on the PRW \cite{zheng2017person} dataset.

\subsection{Datasets and Settings}

\textbf{CUHK-SYSU.} Scene images in the CUHK-SYSU \cite{xiao2017joint} are collected from real street snaps and movies. 
There are a total of 18,184 realistic and uncropped images, 96,143 annotated bounding boxes and 8,432 different identities.
The dataset is partitioned into two parts without overlap.
The training set includes 11,206 images, 55,272 pedestrians, and 5,532 identities. 
The test set contains 6,978 images, 40,871 pedestrians, and 2,900 identities. 
During inference, for each query, the dataset defines a gallery set with different sizes from 50 to 4,000 to evaluate the performance scalability of models. 
Following the previous works, we report the results with the gallery size of 100 if not specified.

\textbf{PRW.} Images in the PRW \cite{zheng2017person} dataset are collected by 6 static cameras at Tsinghua university. 
There are 11,816 video frames and 43,110 annotated bounding boxes. 34,304 of these boxes are annotated with 932 labelled identities and the rest are marked as unknown identities.
It is also divided into two groups.
The training set contains 5,704 images, 18,048 pedestrians, and 482 identities.
The test set has 6,112 images and 2,057 query persons with 450 identities.
During inference, for each query person, the gallery set is the whole test set, \textit{i.e.}, the gallery size is 6,112.

\textbf{Evaluation Metrics.} Following the previous works \cite{xiao2017joint}, we employ Mean Average Precision (mAP) and Cumulative Matching Characteristics (CMC top-K) to evaluate the performance of the person search. 

\subsection{Implementation Details}

We adopt ResNet50 \cite{he2016deep} and transformer-based PVTv2-B2 \cite{wang2022pvt} that are pre-trained on ImageNet \cite{russakovsky2015imagenet} as backbone. 
To train our model, we adopt the AdamW optimizer with a weight decay rate of 0.0001.
The initial learning rate is set to 0.0001 that is warmed up during the first epoch and decreased by a factor of 10 at 19$^{th}$ and 23$^{th}$ epoch, with a total of 24 epochs.
For CUHK-SYSU/PRW, the circular queue size of OIM is set to 5000/500.
During training, we employ a multi-scale training strategy, where the longer side of the image is randomly resized from 400 to 1666.
For inference, we rescale the test images to a fixed size of 1500 $\times$ 900 pixels. 
For our SeqTR with ResNet50 \cite{he2016deep} backbone, we use one NVIDIA GeForce RTX 3090 to run all experiments and batch size set to 2. 
Our SeqTR with PVTv2-B2 \cite{wang2022pvt} backbone is trained on two RTX 3090 GPUs with batch size set to 1 because of the limitation of GPU memory. 

\subsection{Comparison to the State-of-the-arts}

We compare our SeqTR with the state-of-the-arts, including both two-step models \cite{chen2018person,lan2018person,han2019re,dong2020instance,wang2020tcts} and one-step models \cite{xiao2017joint,liu2017neural,chang2018rcaa,xiao2019ian,yan2019learning,munjal2019query,dong2020bi,zhong2020robust,chen2020norm,kim2021prototype,li2021sequential,han2021decoupled,yan2021anchor,yu2022cascade,cao2022pstr}, on two datasets.

\begin{table}[t!]
  \setlength{\abovecaptionskip}{0.cm}
  \centering
  \normalsize
    \begin{center}
    \resizebox{\linewidth}{!}{
        \begin{tabular}{|c|c|c c|c c|}
        \hline
        \multirow{2}*{\textbf{Method}}& \multirow{2}*{\textbf{Backbone}}& \multicolumn{2}{|c|}{\textbf{CUHK-SYSU}}& \multicolumn{2}{|c|}{\textbf{PRW}} \\
        \cline{3-6}
        ~ & ~ &\textbf{mAP(\%)} & \textbf{Top-1(\%)} & \textbf{mAP(\%)} & \textbf{Top-1(\%)}\\
        \hline
        \hline
        \multicolumn{6}{|l|}{\textit{Two-step methods}} \\
        MGTS \cite{chen2018person}  &  VGG16   &  83.0  &  83.7  &  32.6  &   72.1 \\
        CLSA \cite{lan2018person} &  ResNet50  &  87.2  &  88.5  &  38.7  &   65.0 \\
        RDLR \cite{han2019re}  &  ResNet50   & 93.0  &  94.2  &  42.9  &  70.2 \\
        IGPN \cite{dong2020instance} & ResNet50  & 90.3  & 91.4 &\textbf{47.2} & 87.0 \\
        TCTS \cite{wang2020tcts}   & ResNet50 &\textbf{93.9} &\textbf{95.1} & 46.8  &\textbf{87.5}\\
        \hline
        \hline
        \multicolumn{6}{|l|}{\textit{One-step methods with CNNs}} \\
        OIM \cite{xiao2017joint}  &  ResNet50   &  75.5  &  78.7  &  21.3  &  49.4 \\ 
        NPSM \cite{liu2017neural} &  ResNet50   &  77.9  &  81.2  &  24.2  &  53.1 \\
        RCAA \cite{chang2018rcaa} &  ResNet50   &  79.3  &  81.3  &  -     &   - \\
        IAN \cite{xiao2019ian}    &  ResNet50   &  76.3  &  80.1  &  23.0  &  61.9  \\
        CTXGraph \cite{yan2019learning} &  ResNet50   &  84.1  &  86.5  &  33.4  &  73.6  \\
        QEEPS \cite{munjal2019query} &  ResNet50   &  88.9  &  89.1  &  37.1  &  76.7  \\
        BI-Net \cite{dong2020bi}     &  ResNet50  &  90.0  &  90.7  & 45.3  &  81.7 \\
        APNet \cite{zhong2020robust} &  ResNet50  &  88.9  &  89.3  & 41.9  &  81.4 \\
        NAE \cite{chen2020norm}     &  ResNet50   &  91.5  &  92.4  & 43.3  &  80.9 \\
        NAE+ \cite{chen2020norm}    &  ResNet50   &  92.1  &  92.9  & 44.0  &  81.1 \\
        PGSFL \cite{kim2021prototype} & ResNet50  & 90.2  &  91.8  & 42.5  &  83.5  \\
        SeqNet \cite{li2021sequential} & ResNet50 & 93.8  &  94.6  & 46.7  &  83.4  \\
        DMRN \cite{han2021decoupled}&  ResNet50   & 93.2  &  94.2  & 46.9  &  83.3  \\
        AlignPS \cite{yan2021anchor}&  ResNet50   & 93.1  &  93.4  & 45.9   &  81.9  \\
        \multicolumn{6}{|l|}{\textit{One-step methods with transformers}} \\
        COAT \cite{yu2022cascade}& ResNet50  & 94.2 & 94.7 & 53.3 & 87.4  \\
        PSTR \cite{cao2022pstr}  & ResNet50  & 93.5 &95.0 & 49.5  & 87.8 \\
        SeqTR(Ours)              & ResNet50  & 93.4 & 94.1 & 52.0 & 86.5 \\
        PSTR \cite{cao2022pstr}  & PVTv2-B2 &\textbf{95.2} &\textbf{96.2} &56.5 & 89.7\\
        PSTR* \cite{cao2022pstr} & PVTv2-B2 & 94.6 & 95.6 &  57.6  & \textbf{90.1} \\
        SeqTR(Ours)              & PVTv2-B2 & 94.8  & 95.5  &\textbf{59.3}  &89.4  \\
        \hline
        \hline
        COAT \cite{yu2022cascade}+CBGM & ResNet50  & 94.8  &95.2  &54.0  &89.1 \\
        PSTR \cite{cao2022pstr}+CBGM & PVTv2-B2 &\textbf{95.8} &\textbf{96.8}&58.1 &\textbf{92.0} \\
        PSTR* \cite{cao2022pstr}+CBGM & PVTv2-B2 & 95.2 & 96.1 & 58.2 &91.5\\
        SeqTR(Ours)+CBGM  & PVTv2-B2  & 95.4  & 96.3 & \textbf{59.8}  & 90.6 \\
        \hline
        \end{tabular}
        }
    \end{center}
    \caption{Comparison with the state-of-the-art methods on CUHK-SYSU and PRW test sets. * denotes our reproduced result.
    The highest scores in each group are highlighted in bold.}
    \label{Tab:comparison}
    \vspace{-0.2cm}
\end{table}

\textbf{Results on CUHK-SYSU.}
As shown in Table \ref{Tab:comparison}, our SeqTR outperforms most one-step methods and achieves comparable performance to two-step methods on the CUHK-SYSU test set \cite{xiao2017joint}. 

The best two-step method TCTS \cite{wang2020tcts} achieves mAP scores of 93.9\%. Among one-step methods with the ResNet50 \cite{he2016deep}, COAT \cite{yu2022cascade} achieves the best mAP score of 94.2\%. Our SeqTR with the same ResNet50 backbone, which achieves comparable 93.4\% mAP and 94.1\% top-1 accuracy, outperforms AlignPS \cite{yan2021anchor} by 0.3\% and 0.7\% in mAP and top-1 accuracy, respectively. Our results are slightly worse than the transformer-based COAT \cite{yu2022cascade} and PSTR \cite{cao2022pstr}. 

Then, based on PVTv2-B2 \cite{wang2022pvt} backbone, the performance of our SeqTR is significantly improved to 94.8\% mAP and 95.5\% top-1 accuracy. For a fair comparison, we reproduce the performance of PSTR \cite{cao2022pstr} with the same PVTv2-B2 \cite{wang2022pvt} backbone (named PSTR*) to eliminate the effects of different training strategies, \textit{i.e.}, single-GPU training and distributed training. 
Specifically, we set batch size from 2 to 1 and use two RTX 3090 GPUs for distributed training three times. The average of the three reproduced results is then calculated and reported in Table \ref{Tab:comparison}. 
Our method outperforms the reproduced results of PSTR by 0.2\% in mAP.
Moreover, the post-processing strategy Context Bipartite Graph Matching(CBGM) \cite{li2021sequential} is widely used to  improve mAP and top-1 accuracy. By employing CBGM, our SeqTR achieves 95.4\% mAP and 96.3\% top-1 accuracy, which outperforms the  reproduced results of PSTR* with CBGM.

\begin{figure}[t!]
    \setlength{\abovecaptionskip}{0.cm}
    \begin{center}
        \includegraphics[width=0.478\textwidth]{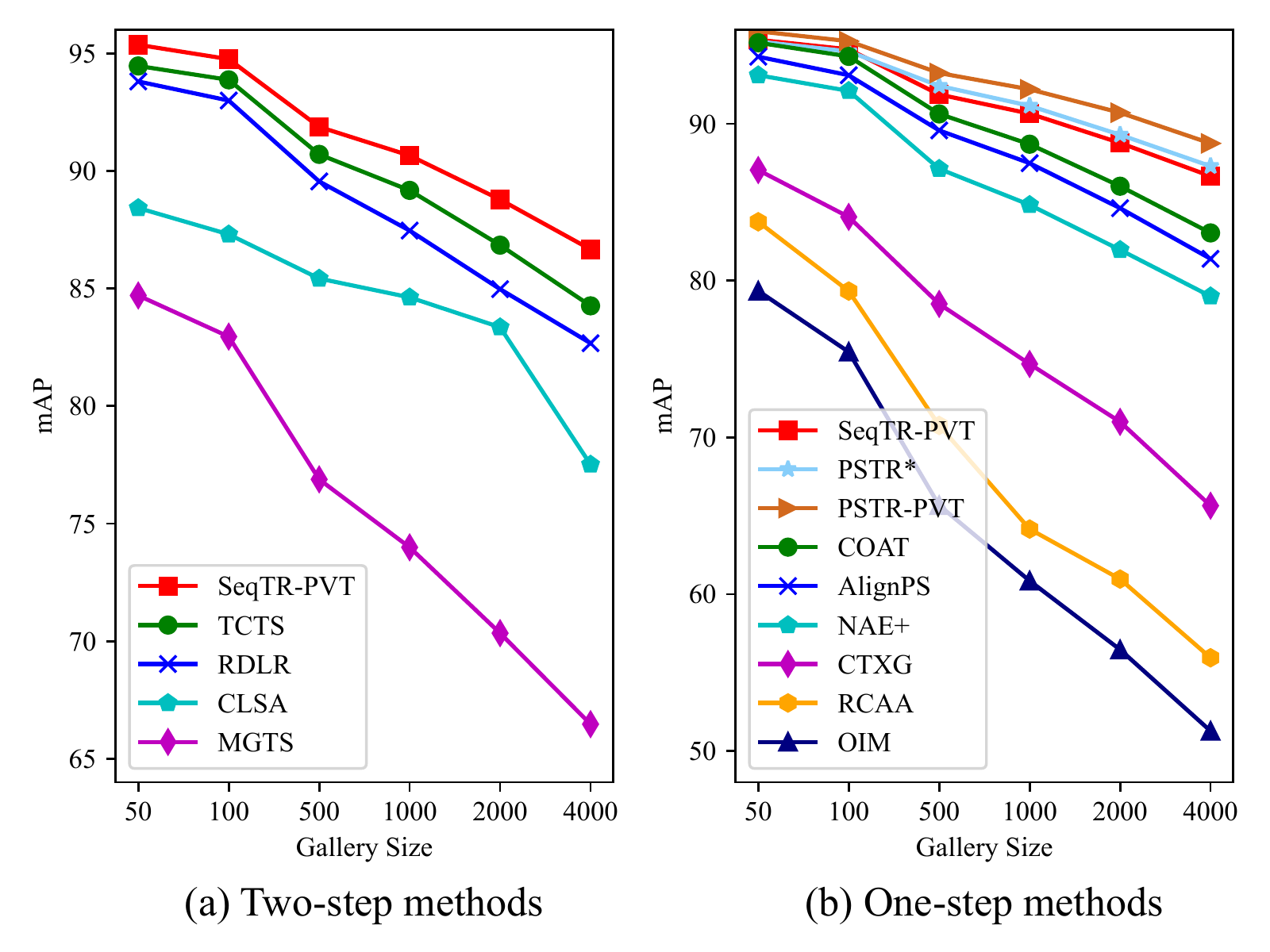}
        \caption{Comparison with (a) two-step models and (b) one-step models on CUHK-SYSU with different gallery sizes.}\label{fig:cuhk}
    \end{center}
\end{figure}

We also evaluate the performance scalability of these models with different gallery sizes. Fig. \ref{fig:cuhk} shows that the mAP of all methods decreases monotonically as the gallery size increases, which illustrates the fact that more distracting persons introduced in the larger gallery make searching much more difficult.
As shown in Fig. \ref{fig:cuhk}, our SeqTR outperforms most models.

\textbf{Results on PRW.}
The PRW dataset \cite{zheng2017person} is more challenging than the CUHK-SYSU dataset \cite{xiao2017joint} for less training data and larger gallery size. Furthermore, there is a large number of people wearing similar uniforms and there are more scale variations, pose/viewpoint changes and occlusions.
Nevertheless, as can be observed from Table \ref{Tab:comparison}, our method achieves strong performance.

With ResNet50 \cite{he2016deep} backbone, our SeqTR achieves 52.0\% mAP and 86.5\% top-1 accuracy, outperforming all two-step methods and with a significant gain of 2.5\% mAP than PSTR \cite{cao2022pstr} with the same backbone.  The performance of our method is slightly lower than COAT \cite{yu2022cascade} by 1.3\% mAP and 0.9\% top-1 accuracy. 

 With PVTv2-B2 backbone \cite{wang2022pvt}, our SeqTR achieves 59.3\% mAP and 89.4\% top-1 accuracy,  outperforming all existing methods with a clear margin on mAP. 
We attribute it to our designed re-ID transformer which  alleviates some challenges, such as recognizing the query person from co-travellers wearing the same uniform. 
Finally, our SeqTR is improved to the best 59.8\% mAP and comparable 90.6\% top-1 accuracy with CBGM.

\subsection{Ablation Study}

We perform a series of ablation studies on the PRW \cite{zheng2017person} dataset to analyze our design decisions. Limited by the memory of the RTX 3090 GPU, we choose the ResNet50 \cite{he2016deep} backbone to void the impact of the distributed training. 

\textbf{Re-ID Transformer Structure.}
Setting the number of the self-attention layer in each transformer layer to 1, we evaluate the impact of the number of transformer layers $M$ and the number of cross-attention layers $K$. As shown in Table \ref{Tab: RDS}, when the number of transformer layers is greater than 2, different combinations of $M$ and $K$ have a slight impact on performance. Among these configurations, when $M = 3$ and $K = 3$, our SeqTR achieves the best performance of 52.0\% mAP and 86.5\% top-1 accuracy.

\begin{table}[t!]
  \setlength{\abovecaptionskip}{0.cm}
  \centering
  \normalsize
    \begin{center}
    \resizebox{\linewidth}{!}{
        \begin{tabular}{|c|c|c c|}
            \hline
            \textbf{Transformer layers}&\textbf{Cross-attention }&\multirow{2}*{\textbf{mAP(\%)}}&\multirow{2}*{\textbf{Top-1(\%)}}\\
           \bm{$M$} & \textbf{layers \bm{$K$}} & & \\
            \hline
            \multirow{3}*{2} &     2      &   50.8  & 86.5 \\ 
                        ~    &     3      &   50.4  &  86.0 \\ 
                        ~    &     4      &   49.9  &  85.5  \\ 
            \hline
            \multirow{3}*{3} &     2      &   50.4  &  86.7  \\ 
                    ~        &     3      &  52.0  &  86.5  \\ 
                    ~        &     4      &   50.7  &  86.8  \\ 
            \hline
            \multirow{3}*{4} &     2      &   50.5  &  86.7  \\ 
                    ~        &     3      &   50.3  &  86.1  \\ 
                    ~        &     4      &   50.6  &  86.7  \\ 
            \hline
        \end{tabular}
        }
    \end{center}
    \caption{Ablation study for different shared re-ID transformer structures on PRW dataset. 
    }
    \label{Tab: RDS}
    \vspace{-0.2cm}
\end{table}

\begin{table}[t!]
  \setlength{\abovecaptionskip}{0.cm}
  \centering
  \normalsize
    \begin{center}
    \resizebox{\linewidth}{!}{
        \begin{tabular}{|c|c c|}
            \hline
            \textbf{Method} & \textbf{mAP(\%)}  & \textbf{Top-1(\%)}\\
            \hline
            re-ID transformer              &   52.0   &  86.5 \\ 
            re-ID transformer w/o self-attention layers&   49.6   &  85.5 \\ 
            \hline
        \end{tabular}
        }
    \end{center}
    \caption{Comparative results of adding and removing self-attention layer on PRW dataset. 
    }
    \label{Tab:SAL}
\end{table}

\textbf{Importance of Self-Attention Layer.}
We also evaluate the importance of the self-attention layer. In Table \ref{Tab:SAL}, we find that adding self-attention layers yields improvements of 2.4\%  on mAP and 1\%  on top-1 accuracy respectively.

\textbf{Schemes of Employing Multi-scale Features.}
To evaluate the effect of different re-ID transformer schemes, we design three different variants as illustrated in Fig. \ref{fig:RDDS} and report the results in Table \ref{Tab:DRD}. 
First, For a multi-scale re-ID transformer-$d$, it outputs $d$ dimensional re-ID feature embeddings for matching. We obtain 44.1\% on mAP and 80.9\% on top-1 accuracy.
To align with the $3d$ dimensional matching embeddings of other schemes (Fig. \ref{fig:RDDS}(b) and Fig. \ref{fig:RDDS}(c)), we also design a multi-scale re-ID transformer-$3d$. However, it has no improvement on mAP.
Compared to the multi-scale re-ID transformer-$3d$, the parallel re-ID transformer (Fig. \ref{fig:RDDS}(b)) has absolute gains of 7.0\% on mAP and 2.1\% on top-1 accuracy.
Then, the shared re-ID transformer (Fig. \ref{fig:RDDS}(c)) achieves the best performance with 52.0\% on mAP and 86.5\% on top-1 accuracy.

\begin{figure*}[t!]
  \setlength{\abovecaptionskip}{0.cm}
  \begin{center}
    \includegraphics[width=1\textwidth]{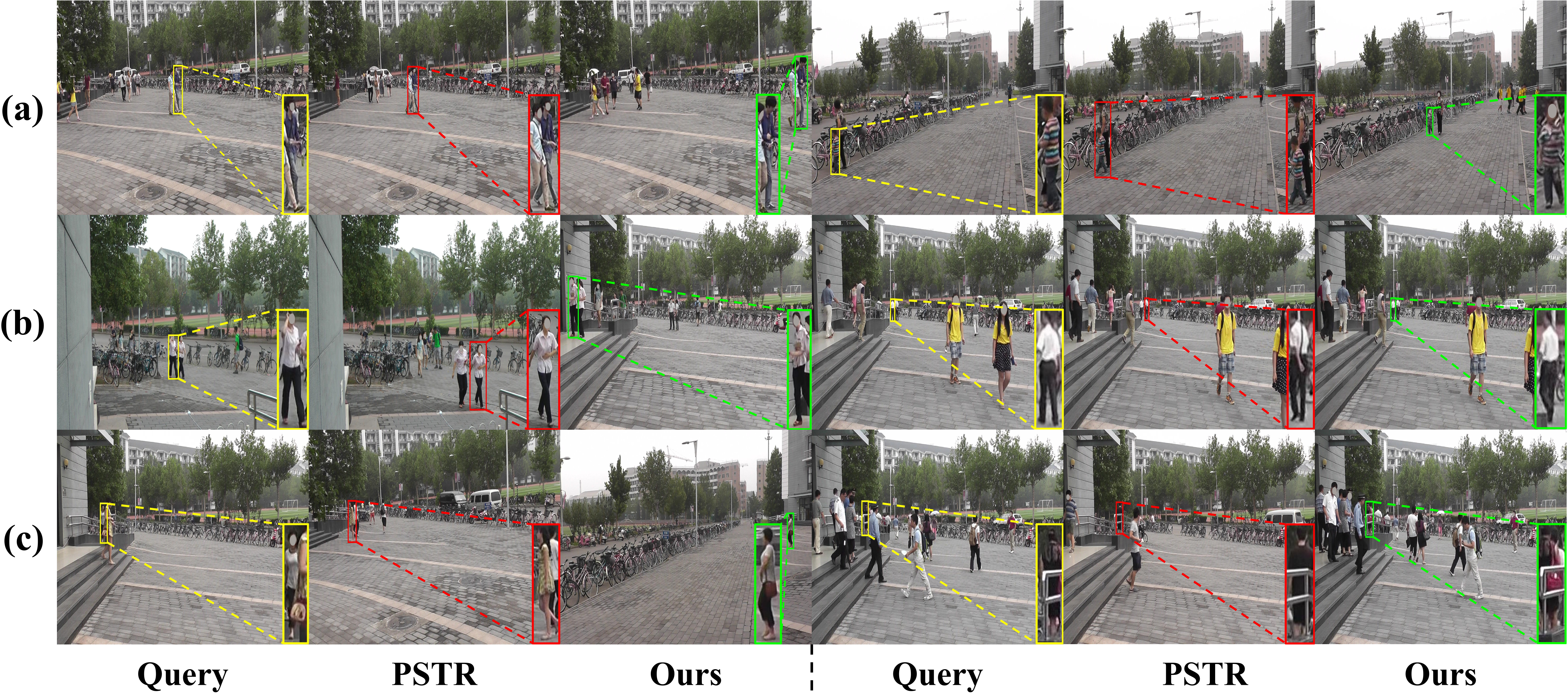}
    \caption{Qualitative comparison with PSTR \cite{cao2022pstr}. The yellow bounding boxes denote the queries, while the green and red bounding boxes denote correct and incorrect top-1 matches, respectively.
    Row (a) are two cases to illustrate the strength of the sequential framework.
    Row (b) are two cases to show the importance of self-attention layers in our re-ID transformer. 
    Row (c) are two cases to show the advantages of cross-attention layers in our re-ID transformer. 
    }\label{fig:vsi}
  \end{center}
  \vspace{-0.3cm}
\end{figure*}

\begin{table}[t!]
  \setlength{\abovecaptionskip}{0.cm}
  \centering
  \normalsize
    \begin{center}
    \resizebox{\linewidth}{!}{
        \begin{tabular}{|c|c c|}
        \hline
        \textbf{Re-ID transformer scheme}&\textbf{mAP(\%)}& \textbf{Top-1(\%)} \\
        \hline
        Multi-scale re-ID transformer-$d$& 44.1 & 80.9  \\
        Multi-scale re-ID transformer-$3d$& 44.1 & 83.1  \\
        Parallel re-ID transformer    & 51.1 & 85.2  \\
        Shared re-ID transformer      & 52.0 & 86.5 \\
        \hline
        \end{tabular}
        }
    \end{center}
    \caption{Comparative results with different variants of the re-ID transformer on PRW dataset.
    }
    \label{Tab:DRD}
\end{table}

\begin{table}[t!]
  \setlength{\abovecaptionskip}{0.cm}
  \centering
  \normalsize
    \begin{center}
    \resizebox{\linewidth}{!}{
        \begin{tabular}{|c|c c c c|c c|}
        \hline
        \textbf{Input feature}&\textbf{E}&\bm{$F_{b4}$}&\bm{$F_{b3}$}&\bm{$F_{b2}$}&\textbf{mAP(\%)}&\textbf{Top-1(\%)}\\
        \hline
        \multirow{4}*{Single-scale feature} &\checkmark &  ~  & ~  & ~ & 26.5 & 66.4 \\
        ~                                   & ~         &\checkmark& ~  & ~ & 41.6 & 79.5 \\
      ~                                   & ~         & ~ &\checkmark& ~ & 45.6 &  82.9 \\
        ~                                   & ~         & ~ & ~ & \checkmark&41.7 & 82.6  \\
        \hline
        \multirow{3}*{Multi-scale feature} & ~ &  \checkmark  &  ~    &  ~   & 41.6  & 79.5 \\
         ~ & ~ &\checkmark &\checkmark & ~   & 47.8   & 82.7 \\
         ~ & ~ &\checkmark &\checkmark &  \checkmark    & 52.0 & 86.5 \\
        \hline
        \end{tabular}
        }
    \end{center}
    \caption{Comparative results by employing different input features on PRW dataset. "\checkmark" means using the corresponding feature. "E" denotes the output feature of the encoder in the detection transformer. 
    }
    \label{Tab:feat}
\end{table}

\textbf{Choices of Input Features to re-ID transformer.}
We conduct experiments on employing different input features to the shared re-ID transformer, including single-level and multi-scale features. The results are reported in Table \ref{Tab:feat}.
 Specifically, we first evaluate the single-level feature respectively. 
Among these single-level features, the output feature of the encoder in the detection transformer provides less information for the re-ID task and is discarded in later experiments. 
Relatively, C4 yields the best performance.
 Furthermore, we also show the performance of utilizing multi-scale features. 
As can be observed,  the best performance is  achieved by using three-level features. 

\textbf{Efficiency Comparison.}
Generally, there are more pedestrians in every scene image in the PRW \cite{zheng2017person} dataset.
To evaluate our contributions in the sequential framework, we conduct runtime efficiency analysis on PRW \cite{zheng2017person} dataset.
As shown in Table \ref{Tab:time}, our SeqTR with ResNet50 \cite{he2016deep} backbone takes 86 milliseconds to process an image, which is faster than SeqNet \cite{li2021sequential} and COAT \cite{yu2022cascade}. It is attributed to the design without requiring an NMS. 
For using PVTv2-B2 backbone \cite{wang2022pvt}, our SeqTR is slower than PSTR \cite{cao2022pstr}, but achieves an absolute of 2.8\% mAP over PSTR \cite{cao2022pstr}. Our SeqTR with PVTv2-B2 backbone has the same speed of 130 milliseconds with COAT \cite{yu2022cascade} with ResNet50 backbone, however our method achieves +6.0\%  and +2.0\% gains of mAP and top-1 accuracy respectively.

\begin{table}[t!]
  \setlength{\abovecaptionskip}{0.cm}
  \centering
  \normalsize
    \begin{center}
    \resizebox{\linewidth}{!}{
        \begin{tabular}{|c|c|c|c|c|c|}
            \hline
            \textbf{Method} & \textbf{Backbone} & \textbf{GPU}  & \textbf{Time(ms)}&\textbf{mAP(\%)}&\textbf{Top-1(\%)}\\
            \hline
            NAE \cite{chen2020norm}        & ResNet50 & RTX 3090 & 80  &  43.3 & 80.9  \\ 
            SeqNet \cite{li2021sequential} & ResNet50 & RTX 3090 & 106 &  46.7 &  83.4 \\
            AlignPS \cite{yan2021anchor}   & ResNet50 & RTX 3090 &\textbf{44} & 45.9 &  81.9 \\
            COAT \cite{yu2022cascade}      & ResNet50 & RTX 3090 & 130 &  53.3 &  87.4 \\
            PSTR \cite{cao2022pstr}        & ResNet50 & RTX 3090 & 52  &  49.5 & 87.8  \\
            SeqTR(Ours)                    & ResNet50 & RTX 3090 & 86  & 52.0  & 86.5 \\
            PSTR \cite{cao2022pstr}        & PVTv2-B2 & RTX 3090 & 88  &  56.5 &\textbf{89.7}\\            
            SeqTR(Ours)                    & PVTv2-B2 & RTX 3090 & 130 &\textbf{59.3}& 89.4 \\
            \hline
        \end{tabular}
        }
    \end{center}
    \caption{Comparative results of person search efficiency on the PRW dataset.}
    \label{Tab:time}
    \vspace{-0.1cm}
\end{table}

\textbf{Qualitative Results.}
To demonstrate the performance of our SeqTR, we show some qualitative comparisons between our SeqTR with PSTR \cite{cao2022pstr} on PRW \cite{zheng2017person} dataset. 
As shown in Fig. \ref{fig:vsi}(a), our SeqTR achieves more accurate pedestrian localizations in both examples, because the sequential framework produces high-quality detection results first that then benefit for the re-ID stage.
In both cases of Fig. \ref{fig:vsi}(b), compared to PSTR \cite{cao2022pstr}, our SeqTR accurately identifies the query persons, whose co-travellers wear similar uniforms. It is attributed to the self-attention layer that employs contextual information. 
In addition, the cross-attention layers in our re-ID transformer contribute to focusing on meaningful regions, although occlusions occur in the given query person in Fig. \ref{fig:vsi}(c). 
The above examples also illustrate that our SeqTR further alleviates some challenges, such as occlusions and distinguishing  similar appearances.

\section{Conclusion}
In this paper, we propose a novel Sequential Transformer (SeqTR) for end-to-end person search. 
Within our SeqTR, a detection transformer and a re-ID transformer are integrated to solve the two contradictory tasks sequentially.
We design a re-ID transformer that contains self-attention layers and cross-attention layers to generate discriminative re-ID feature embeddings. Furthermore, our re-ID transformer adopts a share strategy for employing multi-scale features.
Extensive experiments demonstrate the performance of our proposed framework, which achieves state-of-the-art results on PRW \cite{zheng2017person} dataset.

\clearpage

{
\small
\bibliographystyle{ieee_fullname}
\bibliography{egbib}
}

\end{document}